\documentclass[10pt,twocolumn,letterpaper]{article}

\usepackage[pagenumbers]{cvpr} 

\usepackage[dvipsnames]{xcolor}
\newcommand{\red}[1]{{\color{red}#1}}

\usepackage{multirow}
\usepackage{placeins}

\newcommand{\green}[1]{{\color{green}#1}}

\newcommand{\similar}[2]{\operatorname{sim}{\! \left(#1,#2 \! \right)}}
\newcommand{\avgpool}[2]{\operatorname{avgpool}{(#1,#2 )}}
\newcommand{\concat}[2]{\operatorname{concat}{(#1,#2 )}}
\newcommand{\sg}[1]{\operatorname{sg}{(#1)}}

\newcommand{\expb}[1]{\exp \! \left( #1 \! \right)}

\newcommand{\seq}[3]{\left\{#1\right\}_{#2}^{#3}}
\newcommand{\seqin}[3]{\{#1\}_{#2}^{#3}}

\usepackage{makecell}
\usepackage{algorithm,algpseudocode}

\newcommand{\tabdetsegcoco}[0]{
    \begin{table*}[ht]
        \centering
        \begin{tabular}{l|cccc|ccc|ccc}
            \hline
            \multirow{2}{*}{\textbf{Method}} & \multirow{2}{*}{\textbf{Backbone}} 
            & \multirow{2}{*}{\textbf{\#Epochs}} & \multirow{2}{*}{\textbf{Batch}} 
            & \multirow{2}{*}{\textbf{\#Param (M)}} 
            & \multicolumn{3}{c|}{\textbf{Object Detection}} 
            & \multicolumn{3}{c}{\textbf{Instance Segmentation}} \\
            \cline{6-11}
            &  &  &  &  
            & AP$^{\text{bb}}$ & AP$_{50}^{\text{bb}}$ & AP$_{75}^{\text{bb}}$ 
            & AP$^{\text{mk}}$ & AP$_{50}^{\text{mk}}$ & AP$_{75}^{\text{mk}}$ \\
            \hline
            MoCo-v2~\cite{chen2020improved} & ResNet50 & 800 & 256 & 25.6 
            & 38.5 & 58.1 & 42.1 & 34.8 & 55.3 & 37.3\\
            BYOL \cite{grill2020bootstrap} & ResNet50 & 800 & 256 & 25.6 
            & 39.5 & 59.3 & 43.2 & 35.6 & 56.5 & 38.2\\
            DenseCL \cite{wang2021dense} & ResNet50 & 800 & 256 & 25.6 
            & 39.6 & 59.3 & 43.3 & 35.7 & 56.5 & 38.4\\
            ORL \cite{xie2021unsupervised} & ResNet50 & 800 & 256 & 25.6 
            & 40.3 & 60.2 & 44.4 & 36.3 & 57.5 & 38.9\\
            UniVIP \cite{li2022univip} & ResNet50 & 800 & 256 & 25.6 
            & 40.8 & - & - & 36.8 & - & -\\
            \hline
            MoCo-v3~\cite{chen2021empirical} & ViT-S/16 & 800 & 256 & 22.0 
            & 37.5 & 59.4 & 40.2 & 34.8 & 56.5 & 36.8\\
            DINO~\cite{caron2021emerging}& ViT-S/16 & 800 & 256 & 22.0 
            & 35.3 & 56.8 & 37.6 & 33.3 & 53.8 & 35.1 \\
            iBOT~\cite{zhou2022ibot}& ViT-S/16 & 800 & 256 & 22.0 
            & 38.5 & 60.6 & 41.4 & 35.6 & 57.4 & 37.4 \\
            SelfPatch~\cite{yun2022patch} & ViT-S/16 & 800 & 256 & 22.0 
            & 37.5 & 59.4 & 40.2 & 34.8 & 56.5 & 36.8 \\
            PQCL~\cite{zhang2023patch}& ViT-S/16 & 800 & 256 & 22.0 
            & 38.1 & 59.8 & 41.2 & 35.2 & 56.6 & 37.4 \\
            \textbf{MGC (ours)} & ViT-S/16 & 800 & 256 & 22.0 
            & \textbf{42.0} & \textbf{63.5} & \textbf{45.8}
            & \textbf{38.0} & \textbf{60.2} & \textbf{40.5}\\
            \hline
        \end{tabular}
        \vspace{-1em}
        \caption{The performance comparison on object detection and instance segmentation tasks of MS COCO. 
        All models are \textbf{pretrained on MS COCO2017}, and then finetuned on downstream tasks of MS COCO2017 for $1 \times$ schedule (12 epochs).
        ``\#Epoch'' and ``Batch'' denotes the number of pretraining epochs and the batch size during pretraining. 
        The metrics AP$^{\text{bb}}$ and AP$^{\text{mk}}$ denote bounding box and mask average precision (AP), respectively. 
        }
        \vspace{-2em}
        \label{tab:coco}
    \end{table*}
}

\newcommand{\tabcityscapes}[0]{
    \begin{table}[ht]
        \centering
        \resizebox{\linewidth}{!}{
        \begin{tabular}{l|ccc|ccc}
            \hline
            \multirow{2}{*}{\textbf{Method}} 
            & \multicolumn{3}{c|}{\textbf{obj. det.}} 
            & \multicolumn{3}{c}{\textbf{ins. seg.}} \\
            \cline{2-7}
            & AP$^{\text{bb}}$ & AP$_{50}^{\text{bb}}$ & AP$_{75}^{\text{bb}}$ 
            & AP$^{\text{mk}}$ & AP$_{50}^{\text{mk}}$ & AP$_{75}^{\text{mk}}$\\
            \hline
            MoCo-v3~\cite{chen2021empirical} 
            & 28.5 & 53.5 & 26.5
            & 24.8 & 48.1 & 21.1
            \\
            DINO~\cite{caron2021emerging} 
            & 31.9 & 58.5 & 30.8
            & 27.5 & 51.7 & 24.3
            \\
            iBOT~\cite{zhou2022ibot} 
            & 31.3 & 56.6 & 30.1
            & 27.2 & 50.8 & 24.7
            \\
            SelfPatch~\cite{yun2022patch} 
            & 31.4 & 57.4 & 30.7
            & 27.7 & 51.8 & 26.5
            \\
            PQCL~\cite{zhang2023patch} 
            & 31.1 & 56.5 & 29.9
            & 26.9 & 51.3 & 24.0
            \\
            \textbf{MGC (ours)} 
            & \textbf{33.2} & \textbf{58.7} & \textbf{31.8}
            & \textbf{29.4} & \textbf{54.5} & \textbf{27.3}
            \\
            \hline
        \end{tabular}
        }
        \vspace{-1em}
        \caption{The performance comparison on object detection and instance segmentation tasks of Cityscapes. 
        All ViT-S/16 models are \textbf{pretrained on MS COCO2017}, and then finetuned on downstream tasks of Cityscapes for $1 \times$ schedule (12 epochs).
        }
        \label{tab:cityscapes}
        \vspace{-2em}
    \end{table}
}

\newcommand{\tabade}[0]{
    \begin{table}[ht]
        \centering
        \resizebox{\linewidth}{!}{
        \begin{tabular}{lcccc}
            \toprule
            \textbf{Method} & \textbf{Pretraining} 
            & \textbf{mIoU} & \textbf{aAcc (\%)} & \textbf{mAcc (\%)} \\
            \midrule
            MoCo-v3~\cite{chen2021empirical}& COCO 
            & 33.8 & 76.8 & 43.9 \\
            DINO~\cite{caron2021emerging}& COCO 
            & 27.8 & 74.4 & 37.1 \\
            iBOT~\cite{zhou2022ibot}& COCO 
            & 31.1 & 76.3 & 42.0 \\
            SelfPatch~\cite{yun2022patch}& COCO 
            & 32.9 & 76.4 & 42.8 \\
            PQCL~\cite{zhang2023patch}& COCO 
            & 30.0 & 75.8 & 39.2 \\
            \textbf{MGC (ours)} & COCO 
            & \textbf{37.7} & \textbf{79.0} & \textbf{48.3}\\
            \midrule
            MoCo-v3~\cite{chen2021empirical}& ADE20K 
            & 28.3 & 74.2 & 37.7 \\
            DINO~\cite{caron2021emerging}& ADE20K 
            & 24.1 &  71.8 &  32.6 \\
            iBOT~\cite{zhou2022ibot}& ADE20K 
            & 27.4 & 74.0 &  36.7 \\
            SelfPatch~\cite{yun2022patch}& ADE20K 
            & 25.6 & 72.8 &  34.7 \\
            PQCL~\cite{zhang2023patch}& ADE20K 
            & 30.0 & 75.2 & 39.2 \\
            \textbf{MGC (ours)} & ADE20K 
            & \textbf{31.2} & \textbf{76.2} & \textbf{40.9}\\    
            \bottomrule
        \end{tabular}
        }
        \vspace{-1em}
        \caption{The performance comparison on scene parsing task of ADE20K with ViT-S/16 backbone.
        Each model is \textbf{respectively pretrained on MS COCO2017 and ADE20K}, and then finetuned on ADE20K for 40k iterations. 
        The batch size for pretraining is 256.
        The metrics mIoU, aAcc, and mAcc denote mean intersection of union, all pixel accuracy and mean class accuracy, respectively.
        }
        \vspace{-1em}
        \label{tab:ade}
        \vspace{-1em}
    \end{table}
}

\newcommand{\tabvoc}[0]{
    \begin{table}[t]
        \centering
        \resizebox{\linewidth}{!}{
        \begin{tabular}{lcccc}
            \toprule
            \textbf{Method} & \textbf{Pretraining} 
            & \textbf{mIoU} & \textbf{aAcc (\%)} & \textbf{mAcc (\%)} \\
            \midrule 
            MoCo-v3~\cite{chen2021empirical}& COCO 
            & 56.8 & 89.2 & 67.9 \\
            DINO~\cite{caron2021emerging}& COCO 
            & 43.0 & 85.4 & 55.2 \\
            iBOT~\cite{zhou2022ibot}& COCO 
            & 45.8 & 86.0 & 58.9 \\
            SelfPatch~\cite{yun2022patch}& COCO 
            & 52.6 & 88.5 & 63.6 \\
            PQCL~\cite{zhang2023patch}& COCO 
            & 47.9 & 87.0 & 59.4 \\
            \textbf{MGC (ours)} & COCO 
            & \textbf{64.5} & \textbf{92.0} & \textbf{75.4}\\
            \midrule 
            MoCo-v3~\cite{chen2021empirical}& VOC 
            & 47.0 & 86.8 & 59.0 \\
            DINO~\cite{caron2021emerging}& VOC 
            & 37.9 & 83.7 & 50.0 \\
            iBOT~\cite{zhou2022ibot}& VOC 
            & 44.5 & 85.9 & 57.5 \\
            SelfPatch~\cite{yun2022patch}& VOC 
            & 35.9 & 84.4 & 52.0 \\
            PQCL~\cite{zhang2023patch}& VOC 
            & 44.2 & 85.9 & 56.5\\
            \textbf{MGC (ours)} & VOC 
            & \textbf{54.5} & \textbf{89.2} & \textbf{66.6}\\
            \bottomrule
        \end{tabular}
        }
        \vspace{-1em}
        \caption{The performance comparison on semantic segmentation of PASCAL VOC07+12 with ViT-S/16 backbone.
        Each model is \textbf{respectively pretrained on MS COCO2017 and VOC07+12}, and then finetuned on VOC07+12 for 40k iterations. 
        The batch size for pretraining is 256. 
        ``Pretraining'' denotes the dataset for pretraining.
        }
        \vspace{-1em}
        \label{tab:voc}
    \end{table}
}

\newcommand{\tabkeypoint}[0]{
    \begin{table*}[ht]
        \vspace{-1em}
        \centering
        \begin{tabular}{l|ccc|ccc|ccc}
            \hline
            \textbf{Method}  & \textbf{Backbone} 
            & \textbf{\#Epochs} & \textbf{Batch}
            & \textbf{AP} & \textbf{AP}$\boldsymbol{_{50}}$ & \textbf{AP}$\boldsymbol{_{75}}$ 
            & \textbf{AR} & \textbf{AR}$\boldsymbol{_{50}}$ & \textbf{AR}$\boldsymbol{_{75}}$ 
            \\
            \hline 
            MoCo-v3~\cite{chen2021empirical} & ViT-S/16 & 800 & 256 
            & 69.0 & 88.4 & 76.6 
            & 74.9 & 92.6 & 81.8\\
            DINO~\cite{caron2021emerging} & ViT-S/16 & 800 & 256 
            & 69.5 & 88.5 & 77.6
            & 75.8 & 93.0 & 82.9 \\
            iBOT~\cite{zhou2022ibot} & ViT-S/16 & 800 & 256 
            & 69.0 & 88.4 & 77.0 
            & 75.1 & 93.0 & 82.1 \\
            SelfPatch~\cite{yun2022patch} & ViT-S/16 & 800 & 256 
            & 69.9 & 89.0 & 77.7 
            & 75.8 & 93.2 & 82.8 \\
            PQCL~\cite{zhang2023patch} & ViT-S/16 & 800 & 256 
            & 70.2 & 89.0 & 78.0 
            & 76.7 & 93.2 & 82.8\\
            \textbf{MGC (ours)} & ViT-S/16 & 800 & 256 
            & \textbf{71.6} & \textbf{89.3} & \textbf{79.5} 
            & \textbf{76.9} & \textbf{93.2} & \textbf{84.3}\\
            \hline
        \end{tabular}
        \vspace{-1em}
        \caption{The performance comparison on keypoint detection of MS COCO2017.
        The ViT-S/16 model is \textbf{pretrained on MS COCO2017} and then finetuned on MS COCO2017 for 210 epochs.
        The batch size for pretraining is 256. 
        The metric AR denotes average recall.
        }
        \vspace{-0.5em}
        \label{tab:keypoint}
    \end{table*}
}

\newcommand{\tabablationgrain}[0]{
    \begin{table*}[ht]
        \centering
        \resizebox{\textwidth}{!}{
        \begin{tabular}{cccc|ccc|ccc|ccc|ccc}
            \hline
            \multirow{2}{*}{$\boldsymbol{\mathcal{L}_1}$} 
            & \multirow{2}{*}{$\boldsymbol{\mathcal{L}_2}$} 
            & \multirow{2}{*}{$\boldsymbol{\mathcal{L}_7}$} 
            & \multirow{2}{*}{$\boldsymbol{\mathcal{L}_{14}}$} 
            & \multicolumn{3}{c|}{\textbf{Object Detection}} 
            & \multicolumn{3}{c|}{\textbf{Instance Segmentation}} 
            & \multicolumn{6}{c}{\textbf{Keypoint Detection}}\\
            \cline{5-16}
            &  &  &  
            & AP$^{\text{bb}}$ & AP$_{50}^{\text{bb}}$ & AP$_{75}^{\text{bb}}$ 
            & AP$^{\text{mk}}$ & AP$_{50}^{\text{mk}}$ & AP$_{75}^{\text{mk}}$
            & AP$^{\text{kp}}$ & AP$_{50}^{\text{kp}}$ & AP$_{75}^{\text{kp}}$ 
            & AR$^{\text{kp}}$ & AR$_{50}^{\text{kp}}$ & AR$_{75}^{\text{kp}}$
            \\
            \hline
            \checkmark & -- & -- & -- 
            & 40.1 & 62.1 & 43.4
            & 36.7 & 59.0 & 38.7
            & 70.2 & 88.9 & 78.0
            & 76.1 & 93.1 & 83.0
            \\
            \checkmark & \checkmark & -- & -- 
            & 41.0 & 63.0 & 44.7
            & 37.6 & 60.0 & 39.7
            & 70.3 & 89.0 & 78.2
            & 75.9 & 93.0 & 82.9
            \\
            \checkmark & \checkmark & \checkmark & -- 
            & \textbf{42.0} & \textbf{63.5} & \textbf{45.8} 
            & \textbf{38.0} & \textbf{60.2} & \textbf{40.5} 
            & \textbf{71.6} & \textbf{89.3} & \textbf{79.5} 
            & \textbf{76.9} & \textbf{93.2} & \textbf{84.3} 
            \\
            -- & \checkmark & \checkmark & \checkmark 
            & 40.9 & 63.1 & 44.4
            & 37.4 & 59.8 & 39.7
            & 69.7 & 88.5 & 77.6
            & 75.7 & 92.9 & 82.8
            \\
            \checkmark & \checkmark & \checkmark & \checkmark 
            & 41.0 & 63.0 & 44.6
            & 37.5 & 59.9 & 39.8
            & 70.4 & 88.8 & 78.6
            & 76.2 & 93.1 & 83.4
            \\
            \hline
        \end{tabular}
        }
        \vspace{-1em}
        \caption{The effect of image granularities on object detection, instance segmentation and keypoint detection tasks of MS COCO dataset. 
        }
        \vspace{-1.5em}
        \label{tab:grain}
    \end{table*}
}

\newcommand{\figepoch}[0]{
    \begin{figure*}[ht]
        \centering
        \begin{tabular}{ccc}
            \includegraphics[width=0.3\textwidth]{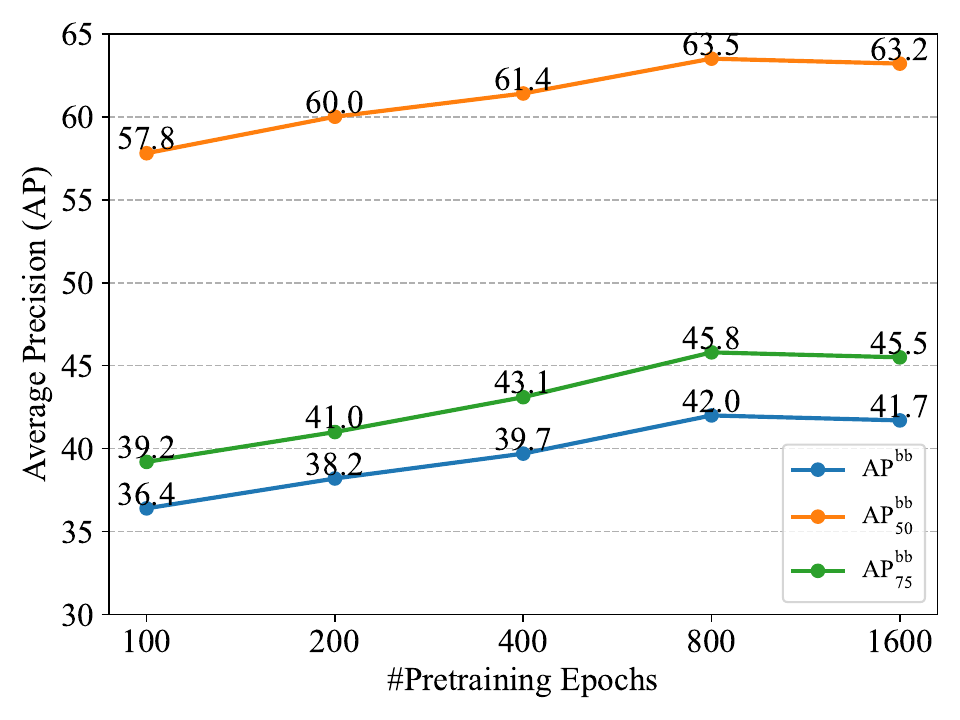}
            & \includegraphics[width=0.3\textwidth]{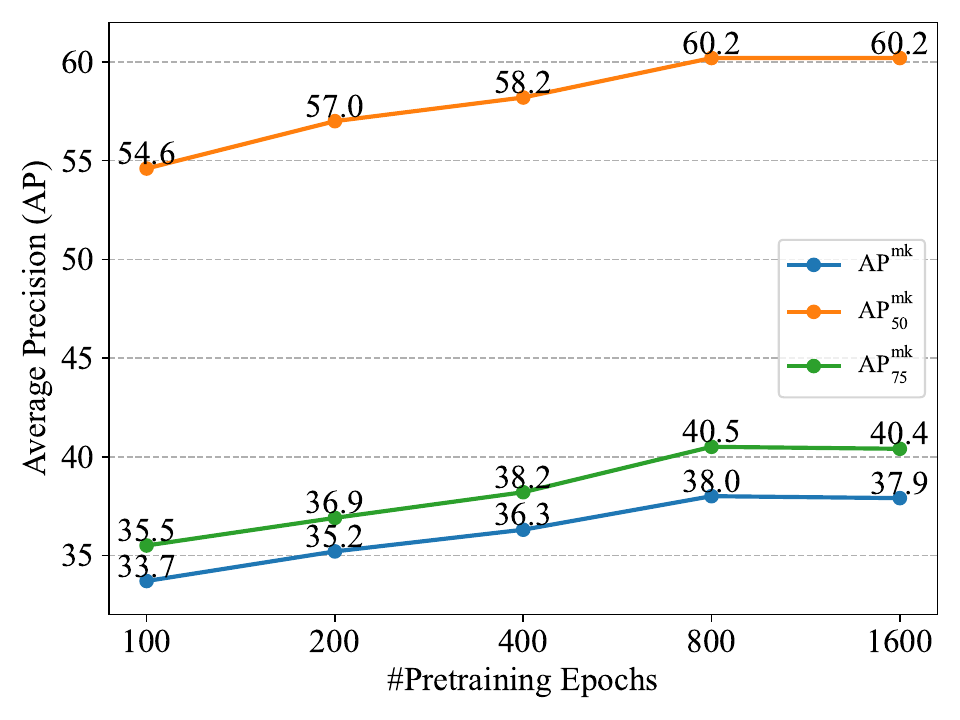}
            & \includegraphics[width=0.3\textwidth]{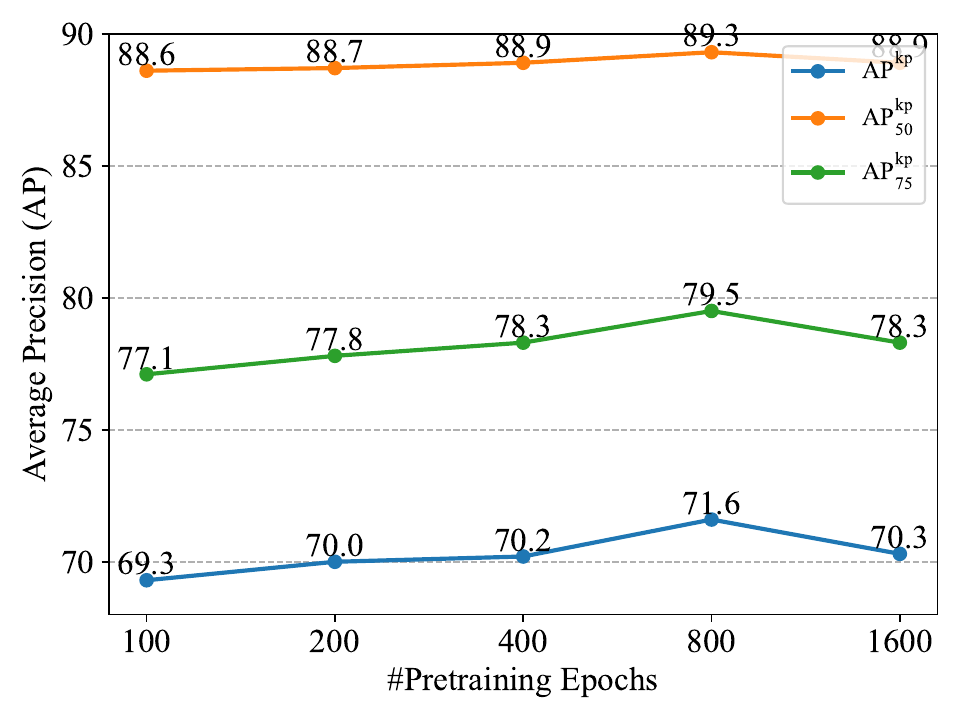}
            \\ 
              {\small (a) object detection task}
            & {\small (b) instance segmentation task}
            & {\small (c) keypoint detection task} \\
        \end{tabular}
        \vspace{-0.8em}
        \caption{The effect of pretraining epochs on object detection, instance segmentation and keypoint detection tasks of COCO dataset.}
        \label{fig:epochs}
        \vspace{-1em}
    \end{figure*}
}

\newcommand{\figcorresp}[0]{
    \begin{figure*}[ht]
        \renewcommand\tabcolsep{1pt}
        \resizebox{\linewidth}{!}
        {
        \begin{tabular}{ccccc}

        \rotatebox{90}{{\scriptsize \quad MGC (SIDV)}} 
        & \includegraphics[width=0.2\linewidth]{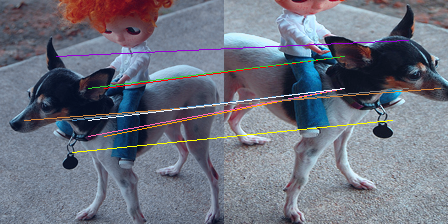} 
        & \includegraphics[width=0.2\linewidth]{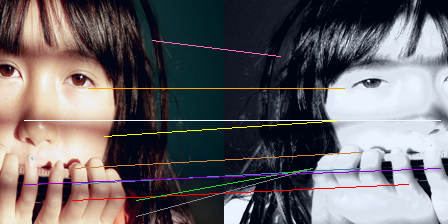} 
        & \includegraphics[width=0.2\linewidth]{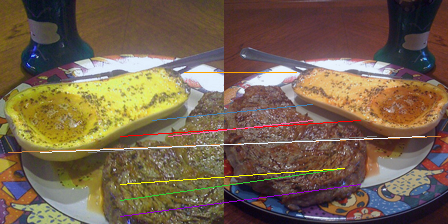} 
        & \includegraphics[width=0.2\linewidth]{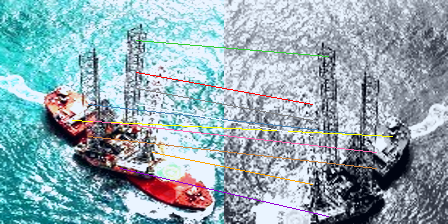} 
        \\
        \rotatebox{90}{{\scriptsize \quad MGC (SCDI)}} 
        & \includegraphics[width=0.2\linewidth]{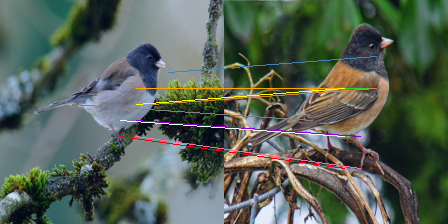} 
        & \includegraphics[width=0.2\linewidth]{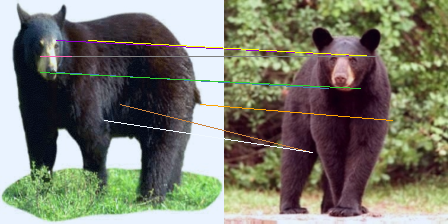} 
        & \includegraphics[width=0.2\linewidth]{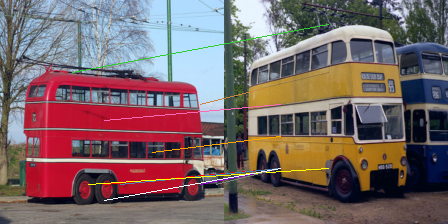} 
        & \includegraphics[width=0.2\linewidth]{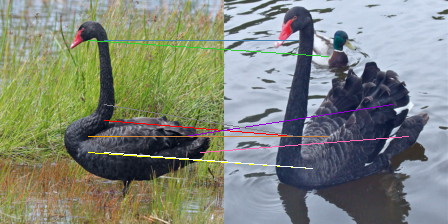} 
        \\
    
        \end{tabular}
        }
        \vspace{-1em}
        \caption{Patch correspondences of our proposed MGC on MS COCO2017. 
        ``SIDV'' refers to the same image but different augmented views. 
        ``SCDI'' refers to the same category but different images. 
        More comparative results can be found in the supplementary material. 
        }
        \label{fig:corresp}
        \vspace{-1em}
    \end{figure*}
}

\definecolor{cvprblue}{rgb}{0.21,0.49,0.74}
\usepackage[pagebackref,breaklinks,colorlinks,citecolor=cvprblue]{hyperref}

\def\paperID{***} 
\def\confName{CVPR}
\def\confYear{2024}

\title{Multi-Grained Contrast for Data-Efficient Unsupervised \\ Representation Learning}

\author{Chengchao Shen\\
Central South University\\
{\tt\small scc.cs@csu.edu.cn}
\and
Jianzhong Chen\\
Central South University\\
{\tt\small cjz\_csu@csu.edu.cn}
\and
Jianxin Wang\\
Central South University\\
{\tt\small jxwang@mail.csu.edu.cn}
}

\begin{document}
\maketitle

\begin{abstract}

    The existing contrastive learning methods mainly focus on single-grained representation learning, e.g., part-level, object-level or scene-level ones, thus inevitably neglecting the transferability of representations on other granularity levels. 
    In this paper, we aim to learn multi-grained representations, which can effectively describe the image on various granularity levels, thus improving generalization on extensive downstream tasks.
    To this end, we propose a novel Multi-Grained Contrast method (MGC) for unsupervised representation learning. Specifically, we construct delicate multi-grained correspondences between positive views and then conduct multi-grained contrast by the correspondences to learn more general unsupervised representations.

    Without pretrained on large-scale dataset, our method significantly outperforms the existing state-of-the-art methods on extensive downstream tasks, including object detection, instance segmentation, scene parsing, semantic segmentation and keypoint detection. 
    Moreover, experimental results support the data-efficient property and excellent representation transferability of our method. 
    The source code and trained weights are available at \url{https://github.com/visresearch/mgc}.
    
\end{abstract}

\section{Introduction}

Contrastive learning~\cite{wu2018unsupervised,ye2019unsupervised,chen2020simple,he2020momentum,chen2021empirical} has achieved remarkable performance on unsupervised visual representation learning. 
The core idea of contrastive learning is to discriminate different augmented views of the same images from other samples, thus encouraging the trained model to capture appearance-invariant representations from a series of augmented images. 
This contrastive configuration works well on single-object-centric datasets, where each image contains a single object as the subject, such as ImageNet~\cite{ILSVRC15}. 
The model pretrained on single-object-centric images is optimized to capture global semantics of the given image, but also easy to miss local semantic patterns, thus degrading the performance on detail-sensitive downstream tasks, such as object detection and instance segmentation. 

To boost the representation performance on detail-sensitive downstream tasks, region-level~\cite{yang2021instance,roh2021spatially,li2022univip,wei2021aligning,xie2021unsupervised,xiao2021region,yun2022patch,zhang2023patch} and pixel-level~\cite{liu2020self,xie2021propagate,pinheiro2020unsupervised,wang2021dense,islam2023self} contrastive learning methods are proposed.
These methods conduct contrastive learning according to the correspondences between regions or pixels from different views and focus on the representations of parts from image, instead of the global ones, thus preserving more details in the representations. 
However, all above methods focus on single semantic granularity, such as object-level semantics or region/pixel-level one, thus unable to achieve optimal performance on extensive downstream tasks.

\begin{figure}[t]
    \centering
    \includegraphics[width=\linewidth]{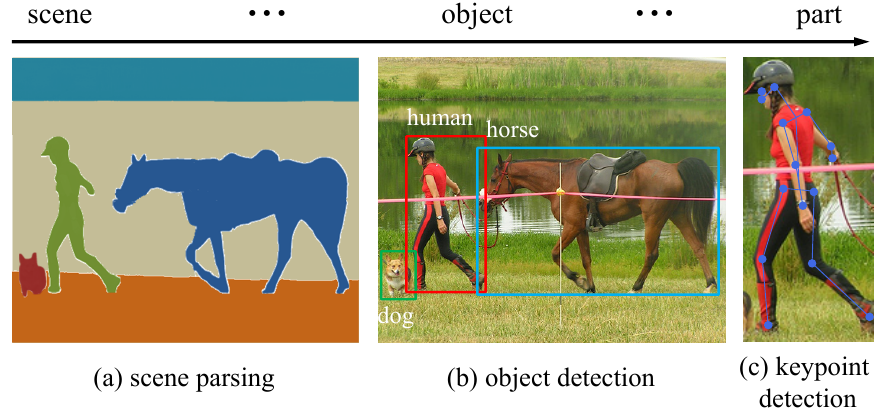}
    \caption{The focus on image granularities of various downstream tasks. 
    For better transferability, general representations are required to cover a wide range of image semantic granularities.
    }
    \vspace{-1em}
    \label{fig:motivation}
\end{figure}

Since different downstream tasks focus on different semantic granularities, unsupervised representations on single granularity can not efficiently be adapted to extensive downstream tasks. 
For example, scene parsing task pays attention to scene-level representations, yet object detection and key point detection respectively focus on object- and part-level ones.
Therefore, for general unsupervised representations of extensive downstream tasks, it's important to model various granularities of image as Figure~\ref{fig:motivation} in the pretext task. 

In this paper, we propose a novel multi-grained contrastive learning method, to model the semantics coherence of positive views on multiple granularities. 
Specifically, we follow the property of vision transformer (ViT)~\cite{dosovitskiy2021vit} to construct general multi-grained correspondences between positive views from patch level to image level. 
For each granularity, we conduct contrastive learning according to delicate correspondence scores between the corresponding regions of positive views.
In this manner, the trained model is guided to localize the region of the given context in another view and capture extensive global and local patterns of images. 
Compared to other contrastive learning methods, our proposed approach provides more accurate correspondences between positive views, thus encouraging the trained model to capture detailed representations on multiple granularities.

Additionally, the existing ViT-based methods suffer a data-hungry dilemma~\cite{touvron2021training,dosovitskiy2021vit}, where the ViT model fails to generalize well without sufficient training data. 
It largely hinders the wide application of the powerful ViT architecture. 
To address this issue, our proposed multi-grained correspondences introduce more diverse and delicate training targets, which significantly improves the data efficiency and the generalization of the trained ViT model. 

Overall, our main contributions can be summarized as follows. 
\begin{itemize}
    \item We propose a novel multi-grained contrast method to represent images on multiple granularities, which effectively improves the transferability of unsupervised representations on extensive downstream tasks. 
    \item Our proposed multi-grained correspondences provide more diverse and elaborate training targets for contrastive learning, thus enabling data-efficient training of ViT without large-scale datasets. 
    \item The proposed method significantly outperforms the previous state-of-the-art methods on extensive popular benchmarks, including MS COCO2017 (object detection, instance segmentation and keypoint detection), ADE20K (scene parsing) and PASCAL VOC07+12 (semantic segmentation) datasets. 
\end{itemize}

\section{Related Work}


\subsection{Image-Level Contrastive Learning}

To learn meaningful representations from unlabeled images, image-level contrastive learning methods~\cite{wu2018unsupervised,ye2019unsupervised} conduct instance discrimination task to discriminate the augmented samples from other image instances. 

Following InstDisc~\cite{wu2018unsupervised} and InvaSpread~\cite{ye2019unsupervised}, SimCLR~\cite{chen2020simple} explores several empirical techniques for contrastive learning, including more and stronger data augmentations, learnable non-linear module, larger batch size and longer training schedule. 
To further improve the stability of contrastive training, momentum encoder~\cite{he2020momentum,chen2020improved,chen2021empirical} is proposed to construct a stable dynamic look-up dictionary.
Additional multi-layer perceptron modules, such as projection head~\cite{chen2020simple,he2020momentum,shen2023asymmetric,shen2023inter} and prediction head~\cite{grill2020bootstrap,chen2021empirical,shen2023asymmetric,shen2023inter}, are introduced to improve the transferability on downstream tasks by decoupling the representations from contrastive pretext task. 

Besides, researchers further study other alternative solutions for contrastive learning. 
BYOL~\cite{grill2020bootstrap} and Simsiam~\cite{chen2021exploring} conduct contrastive learning with only positive pairs and reveal that the negative pair is not the necessary component for contrastive learning. 
SwAV~\cite{caron2020unsupervised} investigates clustering-based contrastive learning method, where the model is trained to align the predictions to clustering codes produced by Sinkhorn-Knopp clustering approach. 
Following SwAV, DINO~\cite{caron2021emerging} replaces Sinkhorn-Knopp clustering method with a unified inner product clustering algorithm on both base and momentum encoder for cluster contrast.

In spite of encouraging performance achieved on unsupervised representation learning, image-level contrastive learning only focuses on image-level representations and inevitably degrades the quality of representations on other granularities, thus weakening the transferability of the corresponding downstream tasks.

\subsection{Region / Pixel-Level Contrastive Learning}

To improve the representation transferability on dense prediction tasks, region- and pixel-level contrastive learning methods are proposed. 

Region-level methods are discussed as follows. 
InsLoc~\cite{yang2021instance} proposes an instance localization task by inserting the foreground objects into background images, where the localization of the target object can be predetermined.
SCRL~\cite{roh2021spatially} and ReSim~\cite{xiao2021region} implement dense representation learning by contrasting the regions from two augmented views. 
SoCo~\cite{wei2021aligning}, ORC~\cite{xie2021unsupervised} and UniVIP~\cite{li2022univip} obtain region correspondences by selective search strategy to perform contrastive learning in the region level.
DetCo~\cite{xie2021detco} introduces additional global views to boost the performance of region-level contrastive learning.
SelfPatch~\cite{yun2022patch} and ADCLR~\cite{zhang2023patch} further explore dense contrastive learning on vision transformer architecture.

Different from the above region-level methods, pixel-level methods conduct contrastive learning on the ``pixels'' of image representations, instead of the regions of original images.
Self-EMD~\cite{liu2020self} regards the Earth Mover's Distance between the features of two positive views as an optimal transportation problem to implement contrastive learning. 
VADeR~\cite{pinheiro2020unsupervised} follows the encoder-decoder architecture adopted on semantic segmentation task to learn contrastive dense representations. 
PixPro~\cite{xie2021propagate}, DenseCL~\cite{wang2021dense} and LCL~\cite{islam2023self} align features of augmented views by contrastive objective in the pixel level.

Nonetheless, the representations obtained from region-level and pixel-level contrastive learning methods are still limited to the representations of single image granularity.
Therefore, these methods can not well address widespread downstream tasks, which require other semantic granularities of images. 
In contrast, our proposed method considers more image granularities, thus effectively improving the representation transferability on various downstream tasks. 
Moreover, compared to the above methods, our method also provides more delicate correspondences between positive views, thus further promoting the quality of unsupervised representations.

\section{Method}

\subsection{Multi-Grained Correspondences}

To learn multi-grained image representations with contrastive learning, we establish multi-grained correspondences between positive pair for contrastive learning. 
In this manner, the learned representations can be effectively transferred to various downstream tasks, each of which requires different semantic granularities. 

The given image $I$ can be augmented into two positive views, $I_1$ and $I_2$, and then patchified into image patch sequences as 
$\seq{\seqin{p_{uv}^{(1)}}{v=0}{V-1}}{u=0}{U-1}$ and 
$\seq{\seqin{p_{uv}^{(2)}}{v=0}{V-1}}{u=0}{U-1}$, 
where $U$ and $V$ are the number of patches along row and column dimension, respectively.
The coordinate of $p_{uv}^{(1)}$ and $p_{uv}^{(2)}$ can be respectively obtained by $(v \cdot w_{1}, u \cdot h_{1}, w_{1}, h_{1})$ and $(v \cdot w_{2}, u \cdot h_{2}, w_{2}, h_{2})$, 
where $u$ and $v$ respectively denote the row and column index of patch; 
$h_1$ and $w_1$ are respectively the height and width of $p_{uv}^{(1)}$; 
$h_2$ and $w_2$ are respectively the height and width of $p_{uv}^{(2)}$.
The coordinate of the overlapping region $I_o$ between $I_1$ and $I_2$ can be presented as $(x_o, y_o, w_o, h_o)$.

\begin{figure}[ht]
    \centering
    \includegraphics[width=\linewidth]{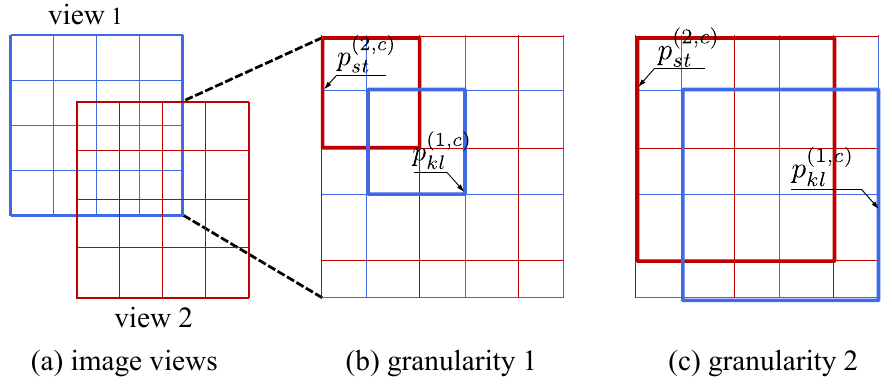}
    \vspace{-2em}
    \caption{Multi-grained correspondences. 
    Adjacent $2 \times 2$ regions of granularity 1 are aggregated into a larger granularity. 
    }
    \label{fig:multigrain}
    \vspace{-1em}
\end{figure}

To obtain larger granularities of images, we apply concatenation operation on adjacent $c \times c$ region of $\seqin{p_{uv}^{(1)}}{u,v}{}$ and $\seqin{p_{uv}^{(2)}}{u,v}{}$ as follows
\begin{small}
\begin{align}
    \seqin{p_{uv}^{(1, c)}}{u,v}{} = \concat{\seqin{p_{uv}^{(1)}}{u,v}{}}{c}, \\
    \seqin{p_{uv}^{(2, c)}}{u,v}{} = \concat{\seqin{p_{uv}^{(2)}}{u,v}{}}{c}, \nonumber
\end{align}
\end{small}
where $c \in \{ 1, 2, 7, 14\}$ denotes the coefficient of granularity or pooling size; 
the size of $p_{uv}^{(1, c)}$ and $p_{uv}^{(2, c)}$ are respectively $w_{1}^{(c)} \times h_{1}^{(c)}$ and $w_{2}^{(c)} \times h_{2}^{(c)}$.

The range of patches in view $I_1$, which overlap with $I_2$ (or $I_o$), can be computed by
\begin{small}
\begin{align}
    k \in \left[ \left\lfloor \frac{y_o - y_1}{h_{1}^{(c)}} \right\rfloor, 
    \left\lceil \frac{y_o + h_o - y_1}{h_{1}^{(c)}} \right\rceil \right], \nonumber \\
    l \in \left[ \left\lfloor \frac{x_o - x_1}{w_{1}^{(c)}} \right\rfloor, 
    \left\lceil \frac{x_o + w_o - x_1}{w_{1}^{(c)}} \right\rceil \right], \nonumber
\end{align}
\end{small}
where $k$ and $l$ are the row and column index of $p_{kl}^{(1, c)}$, respectively.
In the similar way, the range of patches in view $I_2$, which overlap with $p_{kl}^{(1, c)}$, can be obtained by
\begin{footnotesize}
\begin{align}
    s  \in  \left[ 
    \left\lfloor \frac{k \cdot h_{1}^{(c)} + y_1  -  y_2}{h_{2}^{(c)}} \right\rfloor, 
    \left\lceil \frac{(k + 1) \cdot h_{1}^{(c)} + y_1  -  y_2}{h_{2}^{(c)}} \right\rceil 
    \right], \nonumber \\  
    t \in \left[ \left\lfloor \frac{l \cdot w_{1}^{(c)} + x_1 - x_2 }{w_{2}^{(c)}} \right\rfloor, \left\lceil \frac{(l + 1) \cdot w_{1}^{(c)} + x_1 - x_2}{w_{2}^{(c)}} \right\rceil \right],\nonumber
\end{align}
\end{footnotesize}
where $s$ and $t$ are the row and column index of $p_{st}^{(2, c)}$, respectively.

As shown in Figure~\ref{fig:multigrain}, the correspondence score between positive patch pair, $p_{kl}^{(1, c)}$ and $p_{st}^{(2, c)}$, can be defined by their overlapping area as follows
\begin{small}
\begin{equation}
    \setlength\abovedisplayskip{5pt}
    \setlength\belowdisplayskip{5pt}
    w_{klst}^{(c)} = 
    \frac{S(p_{kl}^{(1, c)} \cap p_{st}^{(2, c)})}
    {
        \sum\limits_{s^{\prime},t^{\prime}} 
        S(p_{kl}^{(1, c)} \cap p_{s^{\prime}t^{\prime}}^{(2, c)})
    },
    \footnote{More explanation can be found in the supplementary material.}
\end{equation}
\end{small}
where $S(\cdot)$ denotes the area of the given patch.
In summary, the multi-grained correspondence scores of positive pair can be written as $\seq{ \seqin{\seqin{w_{klst}^{(c)}}{s,t}{}}{k,l}{} }{c}{}$.

\begin{figure*}[ht]
    \centering
    \includegraphics[width=\linewidth]{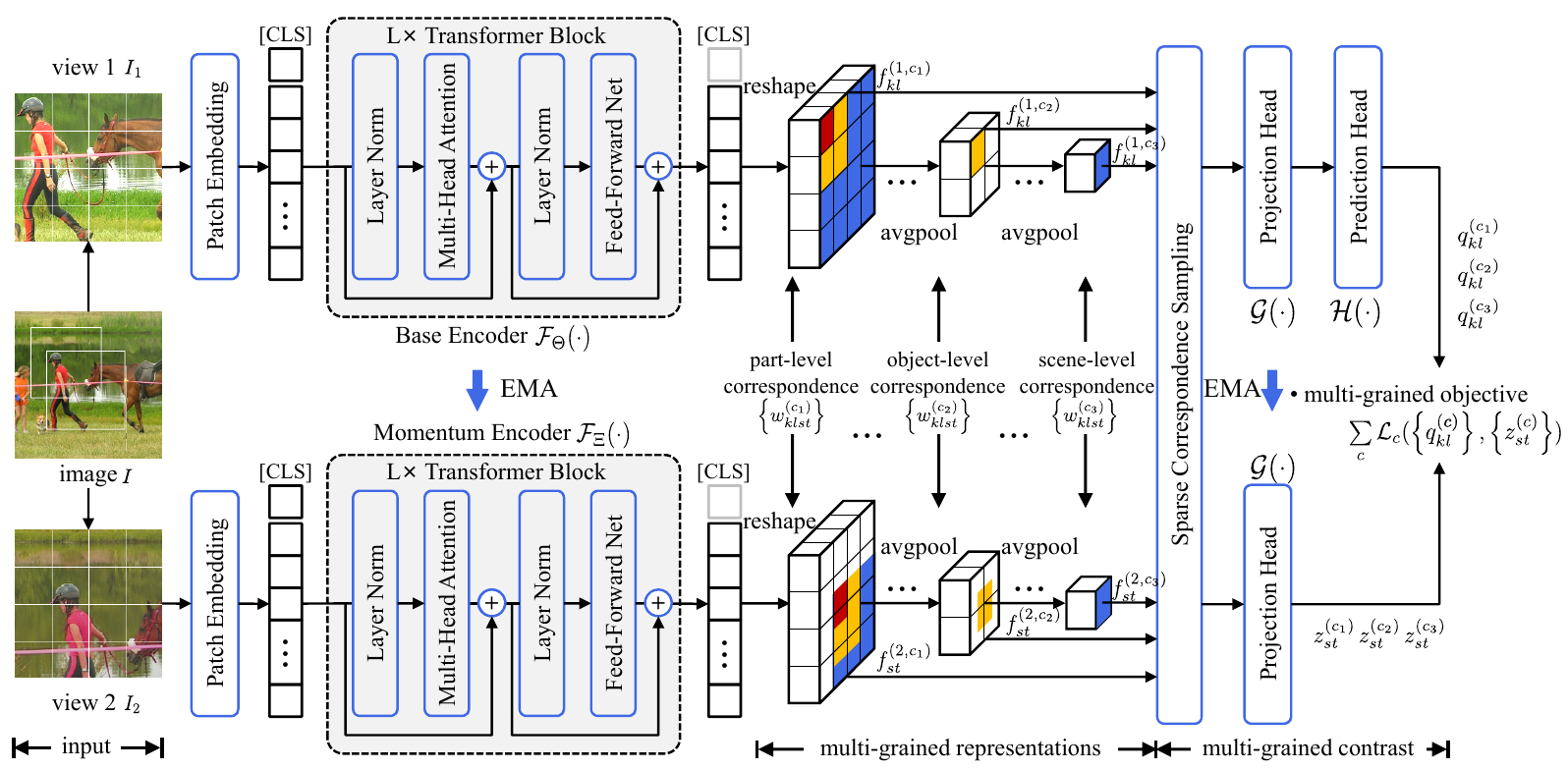}
    \caption{The overview of Multi-Grained Contrast. 
    First, two image views are fed into ViT backbone to obtain patch-wise representations. 
    Then, the representations are aggregated into multi-grained ones and randomly sampled as a sparse sequence to reduce the cost of computation and memory. 
    Finally, the sparse multi-grained representations are optimized by the delicate correspondence targets.
    }
    \label{fig:framework}
    \vspace{-1em}
\end{figure*}

\subsection{Multi-Grained Contrast}

In this section, we apply the above correspondence scores into multi-grained contrastive learning to guide the model to capture extensive semantic granularities of images.

As shown in Figure~\ref{fig:framework}, the image $I$ are augmented by two random data augmentation transforms to obtain $I_1$ and $I_2$. 
To stabilize the training of vision transformer, we introduce additional momentum encoder $\mathcal{F}_{\Xi}$, whose parameters $\Xi$ are updated by the ones of base encoder $\mathcal{F}_{\Theta}$ in an EMA (exponential moving average) manner. 
To improve the transferability of contrastive representations, projection head $\mathcal{G}$ and prediction head $\mathcal{H}$ are also applied.

The image view $I_1$ and $I_2$ are respectively fed into base encoder $\mathcal{F}_{\Theta}$ and momentum encoder $\mathcal{F}_{\Xi}$ to obtain the patch representations 
$\seqin{f_{kl}^{(1)}}{k,l}{} = \mathcal{F}_{\Theta}(I_1)$ and $\seqin{f_{st}^{(2)}}{s,t}{} = \mathcal{F}_{\Xi}(I_2)$. 
Due to the large number of patches, directly feeding all patch representations into MLP-based projection $\mathcal{G}$ or prediction head $\mathcal{H}$ would significantly increase the cost of computation and memory. 
To this end, we conduct a sparse sampling strategy on $\seqin{f_{kl}^{(1)}}{k,l}{}$ and $\seqin{f_{st}^{(2)}}{s,t}{}$ before $\mathcal{G}$ and $\mathcal{H}$ to reduce the cost to a reasonable range. 

To build multi-grained representations, we adopt a series of parameter-free average pooling operations with different window sizes $c$ to aggregate the representations $\seqin{f_{kl}^{(1)}}{k,l}{}$ and $\seqin{f_{st}^{(2)}}{s,t}{}$ as $\seqin{f_{kl}^{(1, c)}}{k,l}{} = \avgpool{\seqin{f_{kl}^{(1)}}{k,l}{}}{c}$ and $\seqin{f_{st}^{(2, c)}}{s,t}{} = \avgpool{\seqin{f_{st}^{(2)}}{s,t}{}}{c}$.

Then, the representation $f_{kl}^{(1,c)}$ from base encoder $\mathcal{F}_{\Theta}$ is sequentially fed into $\mathcal{G}$ and $\mathcal{H}$ to obtain the predicted representation $q_{kl}^{(c)} = \mathcal{H}(\mathcal{G}(f_{kl}^{(1,c)}))$.
Meanwhile, the representation $f_{st}^{(2,c)}$ from momentum encoder $\mathcal{F}_{\Xi}$ is fed into projection head $\mathcal{G}$ to obtain projected representation $z_{st}^{(c)} = \mathcal{G}(f_{st}^{(2,c)})$.

Combining with the above multi-grained representations, we introduce contrastive objective, which measures the cross entropy distance between the predictions and the correspondence targets $w_{klst}^{(i, c)}$ for granularity $c$ as follows,
\begin{footnotesize}
\begin{align}
    &\mathcal{L}_{c} \left(
        \seq{q_{kl}^{(c)}}{k,l}{}, \seq{z_{st}^{(c)}}{s,t}{}
        \right) \\ \nonumber
    &= \!
    - \frac{1}{N} \!
    \sum_{i=0}^{N-1} \!
    \sum_{k,l} \!
    \sum_{s,t} \!
    w_{klst}^{(i, c)} \!
    \log{
        \frac{ \expb{ \similar{ q_{kl}^{(i, c)}}{z_{st}^{(i, c)} }  / \tau } }
        { 
            \sum\limits_{j=0}^{N-1} \!
            \sum\limits_{u, v} \!
            \expb{ 
                \similar{ q_{kl}^{(i, c)}}{z_{uv}^{(j, c)} } 
                / 
                \tau
            } 
        }
    },
\end{align}
\end{footnotesize}
where $N$ denotes batch size and $\tau$ is temperature coefficient.

To guide the trained model to capture multi-grained representations of image, the total objective formed by 4 granularities ($c \in \{ 1, 2, 7, 14\}$) can be written as 
\begin{small}
\begin{equation}
    \mathcal{L}_{\rm total} = 
    \sum_{c \in \{ 1, 2, 7, 14\}} 
    \mathcal{L}_{c}\left(
        \seq{q_{kl}^{(c)}}{k,l}{}, \seq{\sg{z_{st}^{(c)}}}{s,t}{}
    \right),
\end{equation}
\end{small}
where $\sg{\cdot}$ refers to stop gradient operation to alleviate potential representation collapse.

\begin{figure}[ht]
    \centering
    \includegraphics[width=0.8\linewidth]{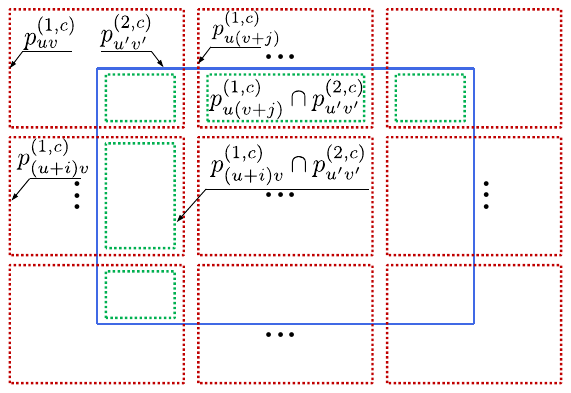}
    \vspace{-1em}
    \caption{Localization analysis. 
    The blue and red boxes refer to the patches from view 1 and view 2, respectively. 
    In this case, the patch localization of view 2 can be obtained by the one of view 1.
    }
    \label{fig:localization}
\end{figure}

\tabdetsegcoco

\subsection{Localization Analysis}

In this section, we verify the multi-grained localization capability of our method, which enables our method to achieve accurate correspondences between positive pair. 
As shown in Figure~\ref{fig:localization}, we find that the location of $p_{uv}^{(1, c)}$ can be computed from the one of $p_{u^{\prime} v^{\prime}}^{(2, c)}$, which overlaps with $p_{uv}^{(1, c)}$ on its edges. 

Without loss of generality, we analyze the process of localization as follows. 
Since the width of $p_{u^{\prime} v^{\prime}}^{(2, c)}$ is split by the overlapping regions $\{ p_{u(v+j)}^{(1, c)} \cap p_{u^{\prime} v^{\prime}}^{(2, c)} \}_j$ and these regions have the same height, the location of $p_{u(v+j)}^{(1, c)}$ along $x$ dimension can be computed by the area ratios of these regions.
Specifically, the area ratios of $p_{u(v+j)}^{(1, c)} \cap p_{u^{\prime} v^{\prime}}^{(2, c)}$ among the overlapping regions can be obtained as follows
\begin{small}
\begin{equation}
    \setlength\abovedisplayskip{5pt}
    \setlength\belowdisplayskip{5pt}
    e_{u(v+j)}^{(w, c)} = 
    \frac{
        S( p_{u(v+j)}^{(1, c)} \cap p_{u^{\prime} v^{\prime}}^{(2, c)} )
    }
    {
        \sum\limits_{k=0}^{N_w - 1} 
        S(p_{u(v+k)}^{(1, c)} \cap p_{u^{\prime} v^{\prime}}^{(2, c)} )
    },
\end{equation}
\end{small}
where $N_w$ denotes the number of patches overlapped with $p_{u^{\prime} v^{\prime}}^{(2, c)}$ along $x$ dimension. 
According to the obtained $e_{u(v+j)}^{(w, c)}$, the $x$ coordinate of $p_{u(v+j)}^{(1, c)} $ can be computed as follows
\begin{small}
\begin{equation}
    \setlength\abovedisplayskip{5pt}
    \setlength\belowdisplayskip{5pt}
    x_j^{(1, c)} = 
    \left(
        v^{\prime} + \sum_{k=0}^{j} e_{u(v+k)}^{(w, c)}
    \right) \cdot w_{2} - w_{1}.
\end{equation}
\end{small}
In the same way, the $y$ coordinate of $p_{(u+i)v}^{(1, c)}$ can be computed as follows
\begin{small}
\begin{align}
    \setlength\abovedisplayskip{5pt}
    \setlength\belowdisplayskip{5pt}
    y_i^{(1, c)} = 
    \left(u^{\prime} + \sum_{k=0}^{i} e_{(u+k)v}^{(h, c)}
    \right) \cdot h_{2} - h_{1}, \\
    e_{(u+i)v}^{(h, c)} = 
    \frac{
        S(p_{(u+i)v}^{(1, c)} \cap p_{u^{\prime} v^{\prime}}^{(2, c)} )
    }
    {
        \sum\limits_{k=0}^{N_h - 1} 
        S(p_{(u+k)v}^{(1, c)} \cap p_{u^{\prime} v^{\prime}}^{(2, c)} )
    }, \nonumber
\end{align}
\end{small}
where $N_h$ denotes the number of patches overlapped with $p_{u^{\prime} v^{\prime}}^{(2, c)}$ along $y$ dimension. 
Therefore, our proposed multi-grained correspondences can provide localization-aware training targets for contrastive learning.

\section{Experiments}

In this section, we evaluate the performance of our proposed method on extensive downstream tasks, including object-, scene- and part-level tasks, to substantiate the effectiveness on multi-grained image representations.

\subsection{Experimental Settings}

\subsubsection{Datasets and Tasks}

To validate the effectiveness of our proposed method, we evaluate the performance of our method on 4 popular benchmark datasets.
Specifically, we respectively pretrain our model on MS COCO2017~\cite{lin2014microsoft}, ADE20K~\cite{zhou2017scene} and PASCAL VOC07+12~\cite{everingham2010pascal} datasets to learn multi-grained representations. 
Then, we evaluate the transferability on various semantic granularities of images by conducting transfer experiments on extensive downstream tasks, including object detection and instance segmentation tasks using Mask R-CNN with FPN~\cite{he2017mask} (on MS COCO2017 dataset and Cityscapes~\cite{cordts2016cityscapes}), scene parsing task using Semantic FPN~\cite{kirillov2019panoptic} on ADE20K, semantic segmentation task using Semantic FPN on PASCAL VOC07+12 and keypoint detection task using ViTPose~\cite{xu2022vitpose} on MS COCO2017 dataset. 
To further verify the effectiveness on large-scale dataset, we also assess our method on ImageNet-1K~\cite{ILSVRC15} and then transfer the pretrained model to object detection and instance segmentation tasks on MS COCO2017 in the supplementary material. 
For all datasets, the image size for pretraining is resized to $224 \times 224$.

\subsubsection{Network and Optimization}
We adopt basic vision transformer~\cite{dosovitskiy2021vit}, ViT-S/16, as the backbone for representation learning. 
The patch size is set to 16 for the experiments on all datasets. 
Following MoCo-v3~\cite{chen2021empirical}, additional MLP-base projection and prediction modules are used to improve the performance of unsupervised representations. 
The structures of projection and prediction modules can be found in the supplementary material. 
The momentum coefficient is set to 0.996 for all experiments.
For the optimization of model pretraining, we adopt AdamW~\cite{loshchilov2017decoupled} optimizer and its learning rate follows a cosine schedule from $10^{-3}$ to $10^{-6}$. 
By default, all models, including all comparative methods, are pretrained for 800 epochs (10 epochs for warmup) with batch size of 256. 
The numbers of sampled tokens for granularities $c \in \{ 1, 2, 7, 14\}$ are 10, 10, 2 and 1, respectively.
Then, the above pretrained models are finetuned on the labeled datasets of the corresponding downstream tasks, to evaluate the representation quality and transferability.
More optimization details for pretraining and finetuning are depicted in the supplementary material. 

\tabcityscapes

\subsection{Experimental Results}

\subsubsection{Object Detection and Instance Segmentation}
To assess the transferability of our pretrained model on object-level tasks, we perform transfer experiments on object detection and instance segmentation tasks of MS COCO2017 dataset using Mask R-CNN with FPN~\cite{he2017mask}. 

As shown in Table~\ref{tab:coco}, the performance of the previous ViT-based contrastive learning methods are obviously inferior to the ResNet-based ones on both object detection and instance segmentation tasks. 
Specifically, both AP$_{75}^{\text{bb}}$ and AP$_{75}^{\text{mk}}$ values of ViT-based methods are significantly lower than the ones of ResNet-based methods, while the AP$_{50}^{\text{bb}}$ and AP$_{50}^{\text{mk}}$ values of ViT-based methods are comparable to the ones of ResNet-based methods. 
For example, iBOT achieves AP$_{75}^{\text{bb}}$ of 41.4 and AP$_{75}^{\text{mk}}$ of 37.4, substantially below the ones of ORL, yet AP$_{50}^{\text{bb}}$ and AP$_{50}^{\text{mk}}$ are comparable to ORL.
It illustrates that these ViT-based methods present inferior representation performance on the small granularities. 

Our proposed ViT-based methods, MGC, achieves AP$^{\text{bb}}$ of 42.0 and AP$^{\text{mk}}$ of 38.0, which significantly outperforms the previous best results of both ViT- and ResNet-based methods. 
It improves the AP$_{50}^{\text{bb}}$ of ViT-S/16 from 60.6 to 63.5, AP$_{75}^{\text{bb}}$ from 41.4 to 45.8, AP$_{50}^{\text{mk}}$ from 57.4 to 60.2, and AP$_{75}^{\text{mk}}$ from 37.4 to 40.5. 
Moreover, the performance gains on both AP$_{75}^{\text{bb}}$ ($4.4 \green{\uparrow}$) and AP$_{75}^{\text{mk}}$ ($3.1 \green{\uparrow}$) are larger than the ones of AP$_{50}^{\text{bb}}$ ($2.9 \green{\uparrow}$) and AP$_{50}^{\text{mk}}$ ($2.8 \green{\uparrow}$).
It demonstrates that our method can effectively improve the quality of object-level representations, especially for the representations of small granularities measured by AP$_{75}^{\text{bb}}$ and AP$_{75}^{\text{mk}}$.

To validate the generality on different datasets, we further conduct transfer experiments on Cityscapes. 
Note that all models are pretrained on MS COCO2017 and then finetuned on Cityscapes for $1 \times$ schedule (12 epochs). 
As Table~\ref{tab:cityscapes}, our method achieves 33.2 AP$^{\text{bb}}$, 58.7 AP$_{50}^{\text{bb}}$ and 31.8 AP$_{75}^{\text{bb}}$ on object detection, outperforming the previous best method DINO by 1.3 AP$^{\text{bb}}$, 0.2 AP$_{50}^{\text{bb}}$ and 1.0 AP$_{75}^{\text{bb}}$, respectively.
Meanwhile, our method consistently outperforms the previous best method on instance segmentation, SelfPatch, by 1.7 AP$^{\text{mk}}$, 2.7 AP$_{50}^{\text{mk}}$ and 0.8 AP$_{75}^{\text{mk}}$.
The above results indicate that our method indeed boosts the representation quality on object-level tasks.

\tabvoc

\tabade

\tabkeypoint

\tabablationgrain

\subsubsection{Semantic Segmentation}

To further verify the effectiveness on object-level representations, we also respectively pretrain our method on MS COCO2017 and PASCAL VOC07+12 datasets, and then finetune the pretrained model using Semantic FPN~\cite{kirillov2019panoptic} on semantic task of PASCAL VOC07+12 for 40k iterations. 

As Table~\ref{tab:voc}, our method with the model pretrained on MS COCO2017 reaches 64.5 mIoU, 92.0\% aAcc and 75.4\% mAcc on semantic segmentation task of PASCAL VOC07+12 and achieves 7.7 mIoU, 2.8\% aAcc and 7.5\% mAcc improvement over the second-best method, MoCo-v3. 
A similar performance improvement is also achieved by the model pretrained on PASCAL VOC07+12.
Our method consistently improves performance over the previous best method, MoCo-v3, from 47.0 to 54.5 mIoU, 86.8\% to 89.2\% aAcc and 59.0\% to 66.6\% mAcc. 

We analyze the insight behind the above results as follows. 
Due to the data hungry property of ViT architecture, the previous unsupervised representation learning methods fail to capture high-quality representations when the size of pretraining dataset, such as MS COCO2017 and PASCAL VOC07+12, is not so large.
The multi-grained training introduced by our method not only improves the efficiency of data utilization, but also the representation quality on various granularities.

\subsubsection{Scene Parsing}

To evaluate the representation transferability on scene-level task: scene parsing, we further conduct comparative experiments about our method and other contrastive methods on ADE20K. 
For more comprehensive comparison, we respectively pretrain the model on MS COCO2017 and ADE20K and then finetune the pretrained model using Semantic FPN~\cite{kirillov2019panoptic} on ADE20K for 40k iterations. 

As shown in Table~\ref{tab:ade}, for the model pretrained on MS COCO2017, our approach achieves 37.7 mIoU, 79.0\% aAcc and 48.3\% mAcc on semantic segmentation task of ADE20K, significantly outperforming the previous state-of-the-art method, MoCo-v3 by 3.9 mIoU, 2.2\% aAcc and 4.4\% mAcc.
When the model pretrained on ADE20K is adopted, our proposed MGC achieves 31.2 mIoU, 76.2\% aAcc and 40.9\% mAcc on semantic segmentation task of ADE20K, surpassing the previous best method PQCL by 1.2 mIoU, 1.0\% aAcc and 1.7\% mAcc, respectively. 
Compared to other contrastive methods, our method achieves significantly superior performance on object-level representations.

\subsubsection{Keypoint Detection}

To validate the representation capability on part-level image granularity, we estimate the transfer performance of our ViT-S/16 model (pretrained on MS COCO2017) on keypoint detection task of MS COCO2017 using ViTPose~\cite{xu2022vitpose}. 
The results are listed in Table~\ref{tab:keypoint}. 
Our method yields 71.6 AP, 89.3 AP$_{50}$, 79.5 AP$_{75}$, 76.9 AR, 93.2 AR$_{50}$ and 84.3 AR$_{75}$, consistently superior to the previous unsupervised representation learning methods. 
The performance gain over the previous state-of-the-art method, PQCL, reaches 1.4 AP, 0.3 AP$_{50}$, 1.5 AP$_{75}$, 0.2 AR and 1.5 AR$_{75}$. 
The above results indeed support that our proposed method also significantly boost the representation quality of smaller part-level granularities. 

\figepoch

\figcorresp

\subsection{Ablation Study}

\subsubsection{The Effect of Image Granularities}

To validate the effectiveness of multiple image granularities, we evaluate the performance of our method under different combinations of image granularities. 
Specifically, the adopted ViT-S/16 model is pretrained on MS COCO2017 for 800 epochs and then finetuned on object detection, instance segmentation and keypoint detection tasks. 

As Table~\ref{tab:grain}, more granularities adopted during pretraining can effectively improve the performance on the above downstream tasks over single-grained baseline. 
When the granularities are increased from $\{ 1 \}$ to $\{ 1,2 \}$, the performance improvements reach 0.9 AP$^{\text{bb}}$ and 0.9 AP$^{\text{mk}}$. 
However, the performance gains on keypoint detection task are not prominent in this case. 
Notably, with the involvement of larger granularities as $\{ 1,2,7,14 \}$, the performance on keypoint detection task is obviously boosted. 
We believe that different granularities can benefit from each other. 
Overall, the results reveal that our proposed method can effectively promote the quality of unsupervised representations.

\subsubsection{The Effect of Pretraining Epochs}

To investigate the effectiveness of the pretraining epochs, we pretrain the ViT-S/16 model on MS COCO2017 with different epochs, and then respectively evaluate the pretrained models on object detection, instance segmentation and keypoint detection tasks of MS COCO2017. 

As shown in Figure~\ref{fig:epochs}, our method can effectively benefit from longer pretraining schedule, especially when the number of pretraining epochs is below 800 epochs. 
Meanwhile, we can observe that our method progressively reaches to the plateau with the increment of the pretraining epochs. 
Especially when the model is pretrained for 1600 epochs, the performance on all downstream tasks tends to slightly decrease. 
Overall, pretraining the model with our method for 800 epochs can achieve the balance between performance and the cost of time.

\subsection{Patch Correspondences}

To validate the localization capability of our method, we analyze the correspondences between different augmented views from the same image and different images with the same category. 
Following iBOT~\cite{zhou2022ibot}, we visualize the correspondences as Figure~\ref{fig:corresp} according to the highest self-attention scores from the last block of ViT-S/16 model, which is pretrained for 800 epochs on MS COCO2017. 
For different views from the same image, the obtained correspondences can well match the corresponding parts from different views, even the appearance variances are large. 
For harder positive pairs from different image with the same category, our method achieves roughly correct matching between the given pairs.
The above results support that our pretrained model possesses the excellent capability of local pattern localization.


\section{Conclusion}

In this paper, we investigate a multi-grained representation issue of unsupervised learning. 
To this end, we propose a novel multi-grained contrast method to align image semantics of positive pairs on various granularity levels. 
The proposed method introduces delicate correspondences on different granularities to encourage the trained model to capture semantic representations on extensive granularities. 
We also give a formal analysis about the multi-grained localization capability of our method. 

Extensive experimental results demonstrate that our method achieves the leading performance on a wide range of downstream tasks, including object detection, instance segmentation, semantic segmentation, scene parsing and keypoint detection.
It indeed supports that our method can effectively model image representations on a wide range of granularities, thus improving the representation transferability on various downstream tasks. 
Moreover, our proposed method presents very promising data-efficient property, where our ViT model generalizes well without pretrained on large-scale dataset. 

\section*{Acknowledgments}
This work was supported by National Key Research and Development Program of China (No.2021YFF1201200), National Natural Science Foundation of China (No.62302523)  and Natural Science Foundation of Hunan Province (No.2022JJ40636). This work was carried out in part using computing resources at the HPC Center of Central South University. 

{
    \small
    \bibliographystyle{ieeenat_fullname}
    \bibliography{main}
}

\begin{equation}
    \setlength\abovedisplayskip{5pt}
    \setlength\belowdisplayskip{5pt}
    w_{klst}^{(c)} = 
    \frac{S(p_{kl}^{(1, c)} \cap p_{st}^{(2, c)})}
    {
        \sum\limits_{s^{\prime},t^{\prime}} 
        S(p_{kl}^{(1, c)} \cap p_{s^{\prime}t^{\prime}}^{(2, c)})
    }.
\end{equation}

\begin{table*}[t]
    \centering
    \vspace{-1em}
    {
        \begin{tabular}{c|c|l|l}
            \hline
            \textbf{Dataset} & \textbf{Layer} & \makecell[c]{\textbf{Projection Head}} & \makecell[c]{\textbf{Prediction Head}} \\
            \hline
            ImageNet-1K & 1 & Linear (2048) + BN + ReLU & Linear (2048) + BN + ReLU\\
                        & 2 & Linear (2048) + BN + ReLU & Linear (128) + BN\textsuperscript{*} \\
                        & 3 & Linear (128) + BN\textsuperscript{*}  & \\
            \hline
        \end{tabular}
    }
    \vspace{-1em}
    \caption{The structure of projection and prediction heads. 
    ``Linear ($m$)'' denotes linear layer with output size $m$.
    ``BN'' and ``ReLU'' denote batch normalization and rectify linear unit operation, respectively. 
    ``BN\textsuperscript{*}'' denotes batch normalization without learnable parameters.
    }
    \label{table:head}
\end{table*}

\begin{table*}[t]
    \centering
    {
       \begin{tabular}{l|l|cc}
          \hline
         \textbf{Augmentation} & \textbf{Parameter} & 
         \textbf{Transform 1} & \textbf{Transform 2}
         \\ 
          \hline
          random crop and resize   & area of the crop                    & [0.08, 1.0] & [0.08, 1.0] \\
                                   & aspect ratio of the crop            & [$3/4$, $4/3$] & [$3/4$, $4/3$] \\
          \hline
          random horizontal flip   & horizontal flip probability         & 0.5 & 0.5 \\
          \hline
          random color jittering   & color jittering probability         & 0.8 & 0.8 \\
                                   & max brightness adjustment intensity & 0.4 & 0.4 \\
                                   & max contrast adjustment intensity   & 0.4 & 0.4 \\
                                   & max saturation adjustment intensity & 0.2 & 0.2 \\
                                   & max hue adjustment intensity        & 0.1 & 0.1 \\
          \hline
          random gray scale        & color dropping probability          & 0.2 & 0.2 \\
          \hline
          random Gaussian blurring & Gaussian blurring probability       & 1.0 & 0.1 \\
                                   & sigma of Gaussian blurring          & [0.1, 2.0] & [0.1, 2.0] \\
          \hline
          random solarization      & solarization probability            & 0.0 & 0.2 \\
          \hline
       \end{tabular}
    }
    \vspace{-1em}
    \caption{The parameters of data augmentations applied during pretraining.
    ``[·, ·]'' denotes the range for uniform sampling.
    }
    \vspace{-1em}
    \label{table:augmentation}
\end{table*}

\section{Multi-Grained Correspondences}

To measure the similarity scores between the overlapped patch pairs, we simply define the training targets by the ratio of overlapping area as follows
\begin{equation}
    s_{klst}^{(c)} = 
    \frac{S(p_{kl}^{(1, c)} \cap p_{st}^{(2, c)})}
    {
        S(p_{kl}^{(1, c)}) + S(p_{st}^{(2, c)})
    }.
\end{equation}
The above relative metric can well address the similarity ranking from the patches with different sizes in the original image. 
To normalize the similarity scores between different patch pairs, we compute the correspondence targets as 
\begin{small}
\begin{equation}
    w_{klst}^{(c)} = 
    \frac{s_{klst}^{(c)}}
    {
        \sum\limits_{s^{\prime},t^{\prime}} 
        {s_{kls^{\prime}t^{\prime}}^{(c)}}
    }
    = \frac{
        S(p_{kl}^{(1, c)} \cap p_{st}^{(2, c)}) / (S(p_{kl}^{(1, c)}) + S(p_{st}^{(2, c)}))
    }{
        \sum\limits_{s^{\prime},t^{\prime}} 
        {S(p_{kl}^{(1, c)} \cap p_{s^{\prime}t^{\prime}}^{(2, c)}) / 
        (S(p_{kl}^{(1, c)}) + S(p_{s^{\prime}t^{\prime}}^{(2, c)}))}
    }.
    \label{eq:original_corresp}
\end{equation}
\end{small}
Due to the constant size of patch $p_{s^{\prime}t^{\prime}}^{(2, c)}$, the Eq.~\ref{eq:original_corresp} can be simplified as

\renewcommand{\algorithmicrequire}{\textbf{Input:}} 
\renewcommand{\algorithmicensure}{\textbf{Output:}}
\begin{algorithm*}[t]
    \caption{Multi-Grained Contrast}
    \label{alg}
    \begin{algorithmic}[1]
        \Require{Unlabeled training data $\mathcal{X} = \{x\}$.}
        \Ensure{The parameters of ViT model $\mathcal{F}(\cdot)$}.
        \State Initialize the parameters of ViT model $\mathcal{F}_{\Theta}$, momentum encoder $\mathcal{F}_{\Xi}$, projector $\mathcal{G}$ and predictor $\mathcal{H}$;
        \For{number of training iterations}
            \State Augment image $I$ to obtain views $I_1$ and $I_2$;
            \State Feed $I_1$ and $I_2$ into ViT model $\mathcal{F}_{\Theta}$ and momentum encoder $\mathcal{F}_{\Xi}$ to obtain $\seqin{f_{kl}^{(1)}}{k,l}{}$ and $\seqin{f_{st}^{(2)}}{s,t}{}$, respectively;
            \State Conduct sparse correspondence sampling on $\seqin{f_{kl}^{(1)}}{k,l}{}$ and $\seqin{f_{st}^{(2)}}{s,t}{}$;
            \State Aggreate the representations $\seqin{f_{kl}^{(1)}}{k,l}{}$ and $\seqin{f_{st}^{(2)}}{s,t}{}$ as $\seqin{f_{kl}^{(1, c)}}{k,l}{}$ and $\seqin{f_{st}^{(2, c)}}{s,t}{}$ with granularity coefficient $c$;
            \State Feed $\seqin{f_{kl}^{(1,c)}}{k,l}{}$ into projector $\mathcal{G}$ and predictor $\mathcal{H}$ to obtain $\seqin{q_{kl}^{(c)}}{k,l}{}$;
            \State Feed $\seqin{f_{st}^{(2,c)}}{k,l}{}$ into projector $\mathcal{G}$ to obtain $\seqin{z_{st}^{(c)}}{k,l}{}$;
            \State Compute the total contrastive objective $\mathcal{L}_{\rm total}$ by Eq.~\red{6}.
            \State Update the parameters of $\mathcal{F}_{\Theta}$, $\mathcal{F}_{\Xi}$, $\mathcal{G}$ and $\mathcal{H}$ using AdamW optimizer.

        \EndFor
    \end{algorithmic}
\end{algorithm*}

\section{Network Structures of Projection and Prediction Modules}

The structures of projection head and prediction head are presented as Table~\ref{table:head}. 
There are 3 linear layers in projection head and 2 linear layers in prediction heads.
The first two linear layers are followed by batch normalization and rectify linear unit in turn, and both output sizes of them are 2048. 
The last linear of both heads are followed by only batch normalization, and output sizes of them are 128.

\section{Data Augmentations}

As shown in Table~\ref{table:augmentation}, we describe the parameters of data augmentations used during self-supervised pretraining. 
For all pretrained datasets, 6 data augmentation techniques are applied to the input images, including random crop and resize, horizontal flip, color jittering, gray scale, Gaussian blurring, as well as solarization. 
There are two differences between the augmentations for the construction of positive pair.
First, the probability of Gaussian blurring is 1.0 in transform 1, but 0.1 in transform 2.
Second, solarization is only used in transform 2.

\section{Algorithm}

To clearly clarify our proposed method, we give the overall algorithm as shown in Algorithm~\ref{alg}.

\begin{table*}[t]
    \centering
    \begin{tabular}{l|cccc|ccc|ccc}
        \hline
        \multirow{2}{*}{\textbf{Method}} & \multirow{2}{*}{\textbf{Backbone}} 
        & \multirow{2}{*}{\textbf{\#Epochs}} & \multirow{2}{*}{\textbf{Batch}} 
        & \multirow{2}{*}{\textbf{\#Param (M)}} 
        & \multicolumn{3}{c|}{\textbf{Object Detection}} 
        & \multicolumn{3}{c}{\textbf{Instance Segmentation}} \\
        \cline{6-11}
        &  &  &  &  
        & AP$^{\text{bb}}$ & AP$_{50}^{\text{bb}}$ & AP$_{75}^{\text{bb}}$ 
        & AP$^{\text{mk}}$ & AP$_{50}^{\text{mk}}$ & AP$_{75}^{\text{mk}}$ \\
        \hline
        MoCo-v2~\cite{chen2020improved} & ResNet50 & 200 & 256 & 25.6 
        & 38.9 & 59.2 & 42.4 & 35.5 & 56.2 & 37.8\\
        BYOL~\cite{grill2020bootstrap} & ResNet50 & 200 & 256 & 25.6 
        & 38.5 & 60.4 & 41.4 & 35.4 & 57.0 & 37.7\\
        DenseCL~\cite{wang2021dense} & ResNet50 & 200 & 256 & 25.6 
        & 40.3 & 59.9 & 44.3 & 36.4 & 57.0 & 39.2\\
        ReSim~\cite{xiao2021region} & ResNet50 & 200 & 256 & 25.6 
        & 40.3 & 60.6 & 44.2 & 36.4 & 57.5 & 38.9\\
        DetCo~\cite{xie2021detco} & ResNet50 & 200 & 256 & 25.6 
        & 40.1 & 61.0 & 43.9 & 36.4 & 58.0 & 38.9\\
        \hline
        MoCo-v3~\cite{chen2021empirical} & ViT-S/16 & 300 & 256 & 22.0 
        & 39.8 & 62.6 & 43.1 & 37.1 & 59.6 & 39.2\\
        DINO~\cite{caron2021emerging}& ViT-S/16 & 300 & 256 & 22.0 
        & 40.8 & 63.4 & 44.2 & 37.3 & 59.9 & 39.5 \\
        iBOT~\cite{zhou2022ibot}& ViT-S/16 & 200 & 256 & 22.0 
        & 42.6 & 65.7 & 47.0 & 39.0 & 61.7 & 41.3 \\
        SelfPatch~\cite{yun2022patch} & ViT-S/16 & 200 & 256 & 22.0 
        & 42.1 & 64.9 & 46.1 & 38.5 & 61.3 & 40.8 \\
        PQCL~\cite{zhang2023patch}& ViT-S/16 & 200 & 256 & 22.0 
        & 43.1 & 66.0 & 47.4 & 39.3 & 62.2 & 41.6 \\
        \textbf{MGC (ours)} & ViT-S/16 & 200 & 256 & 22.0 
        & \textbf{43.3} & \textbf{65.0} & \textbf{47.2}
        & \textbf{39.1} & \textbf{61.8} & \textbf{41.8}\\
        \hline
    \end{tabular}
    \vspace{-1em}
    \caption{The performance comparison on object detection and instance segmentation tasks of MS COCO. 
    All models are \textbf{pretrained on ImageNet-1K}, and then finetuned on downstream tasks of MS COCO2017 for $1 \times$ schedule (12 epochs).
    ``\#Epoch'' and ``Batch'' denotes the number of pretraining epochs and the batch size during pretraining. 
    The metrics AP$^{\text{bb}}$ and AP$^{\text{mk}}$ denote bounding box and mask average precision (AP), respectively. 
    }
    \vspace{-1em}
    \label{tab:in1k}
\end{table*}

\section{More Experimental Settings}

In all experiments, we adopt ViT-S/16 architecture as the backbone. 
The settings of pretraining experiments are clarified as follows. 
For pretraining on MS COCO2017 dataset, the model is trained for 800 epochs using AdamW optimizer and its learning rate follows cosine schedule from $10^{-3}$ to $10^{-6}$. 
During the first 10 pretraining epochs, the learning rate linearly increases from 0 to $10^{-3}$ for warmup. 
The weight decay is set to 0.05. 
For pretraining on ImageNet-1K, the model is trained for 200 epochs using AdamW optimizer and the learning rate follows cosine schedule from $10^{-3}$ to $10^{-6}$. 
The number of linear warmup epoch is set to 30. 
The weight decay is 0.05. 
Drop path regularization is adopted and its coefficient is 0.05.

The settings for finetuning on downstream tasks can be summarized as follows.
For finetuning on object detection and instance segmentation tasks of both MS COCO2017, we train the model for $1\times$ schedule (12 epochs), where the learning rate is $7.5 \times 10^{-3}$ and weight decay is set to $0.03$.
For finetuning on object detection and instance segmentation tasks of Cityscape, we train the model with $769 \times 769$ image resolution for $1\times$ schedule (12 epochs), where the learning rate is $7.5 \times 10^{-3}$ and weight decay is set to $0.03$.
For finetuning on scene parsing task of ADE20K dataset, we train the model with $512 \times 512$ image size for 40k iterations, where the learning rate is $2 \times 10^{-4}$ and weight decay is set to $0.05$.
For finetuning on semantic segmentation task of Pascal VOC07+12 dataset, we train the model  with $512 \times 512$ image size for 40k iterations, where the learning rate is $2 \times 10^{-4}$ and weight decay is set to $0.05$. 
For fineuning on keypoint detection task of MS COCO2017 dataset, we train ViT-S/16 model with $256 \time 192$ input resolution for 210 epochs. 
The learning rate is divided by 10 at the 170 and 200 epochs. 
The layer-wise learning rate decay and stochastic drop path ratio are set to 0.75 and 0.3, respectively.

\section{The Results of The Model pretrained on ImageNet-1K}

To validate the effectiveness of our method on large-scale dataset, we pretrain our model on ImageNet-1K for 300 epochs and then finetune the pretrained model on MS COCO2017 for $1\times$ schedule. 
As shown in Table~\ref{tab:in1k}, with more pretraining data, the performance of all ViT-S/16 model are further improved, especially for other comparative methods. 
It indeed supports the data efficient property of our method. 
In this case, our method consistently achieves significant performance improvement, 0.7 AP$^{\text{bb}}$, 0.5 AP$_{50}^{\text{bb}}$, 0.9 AP$_{75}^{\text{bb}}$, 0.6 AP$^{\text{mk}}$, 0.4 AP$_{50}^{\text{mk}}$ and 0.9 AP$_{75}^{\text{mk}}$ gain over the previous best method, PQCL. 
The experimental results demonstrate that our proposed method also achieves state-of-the-art performance when pretrained on large-scale dataset, ImageNet-1K. 

\begin{figure*}[ht]
    \renewcommand\tabcolsep{1pt}
    \resizebox{\linewidth}{!}
    {
    \begin{tabular}{cccc}

      \includegraphics[width=0.2\linewidth]{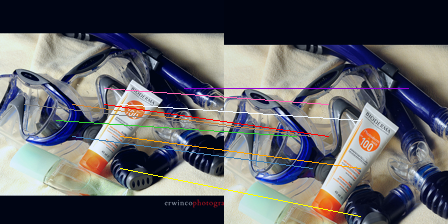} 
    & \includegraphics[width=0.2\linewidth]{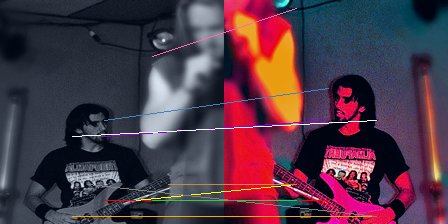} 
    & \includegraphics[width=0.2\linewidth]{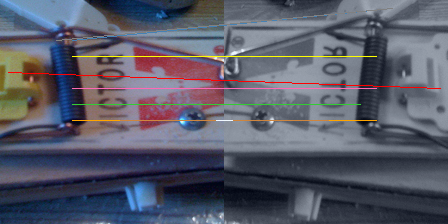} 
    & \includegraphics[width=0.2\linewidth]{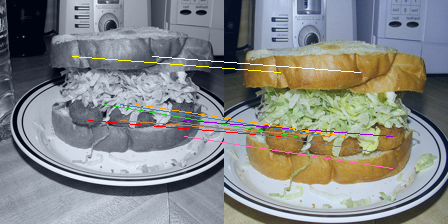} 
    \\
      \includegraphics[width=0.2\linewidth]{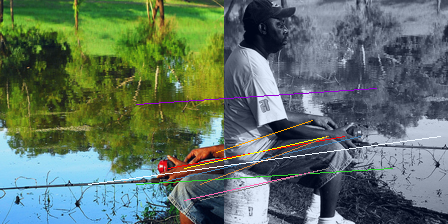} 
    & \includegraphics[width=0.2\linewidth]{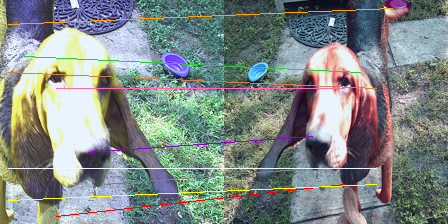} 
    & \includegraphics[width=0.2\linewidth]{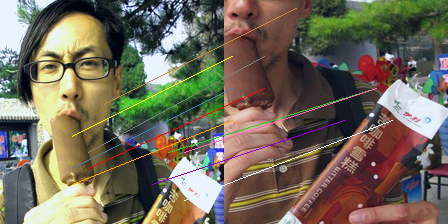} 
    & \includegraphics[width=0.2\linewidth]{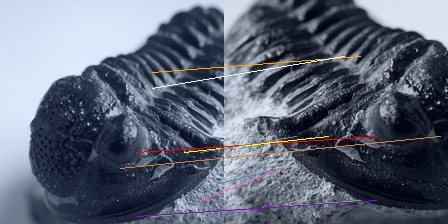} 
    \\
      \includegraphics[width=0.2\linewidth]{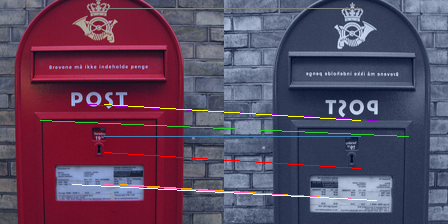} 
    & \includegraphics[width=0.2\linewidth]{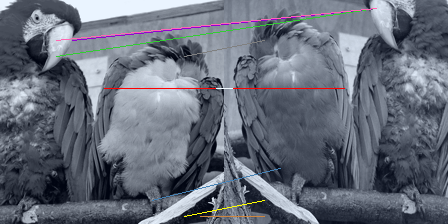} 
    & \includegraphics[width=0.2\linewidth]{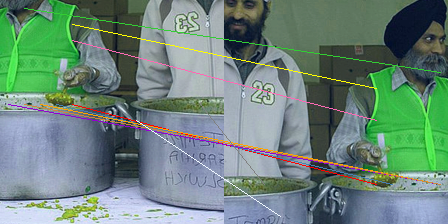} 
    & \includegraphics[width=0.2\linewidth]{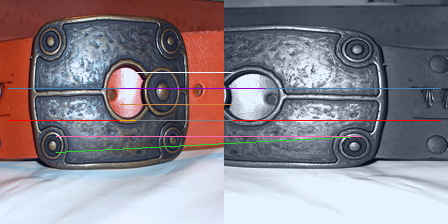} 
    \\
      \includegraphics[width=0.2\linewidth]{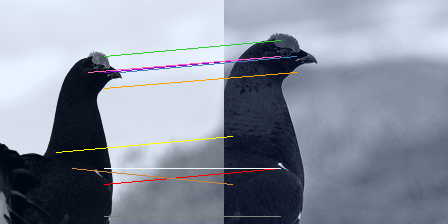} 
    & \includegraphics[width=0.2\linewidth]{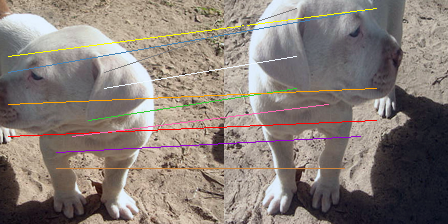} 
    & \includegraphics[width=0.2\linewidth]{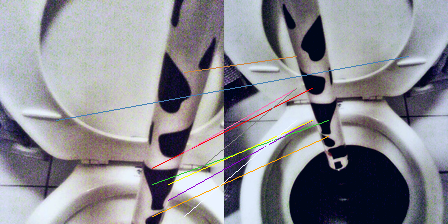} 
    & \includegraphics[width=0.2\linewidth]{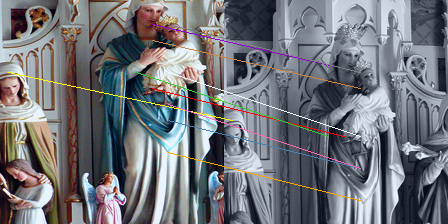} 
    \\

    \end{tabular}
    }
    \vspace{-1em}
    \caption{Patch correspondences of our proposed MGC on MS COCO2017. 
    The given pairs are composed of different augmented views from the same image.
    }
    \vspace{-1em}
    \label{fig:corresp_instance}
\end{figure*}

\begin{figure*}[ht]
    \renewcommand\tabcolsep{1pt}
    \resizebox{\linewidth}{!}
    {
    \begin{tabular}{cccc}

      \includegraphics[width=0.2\linewidth]{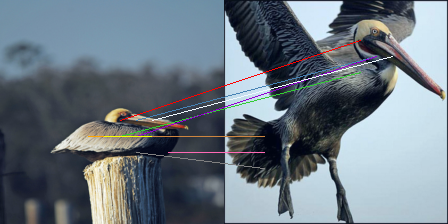} 
    & \includegraphics[width=0.2\linewidth]{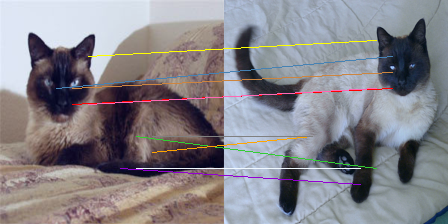} 
    & \includegraphics[width=0.2\linewidth]{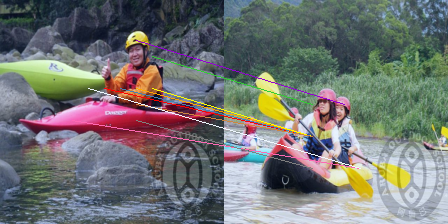} 
    & \includegraphics[width=0.2\linewidth]{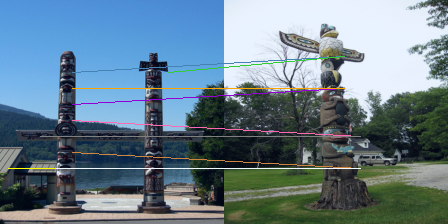} 
    \\
      \includegraphics[width=0.2\linewidth]{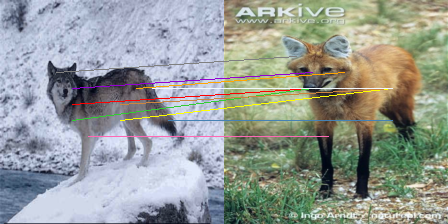} 
    & \includegraphics[width=0.2\linewidth]{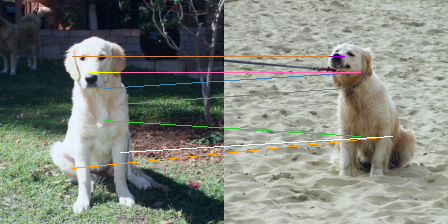} 
    & \includegraphics[width=0.2\linewidth]{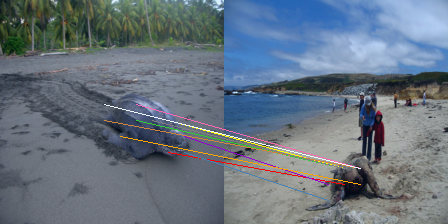} 
    & \includegraphics[width=0.2\linewidth]{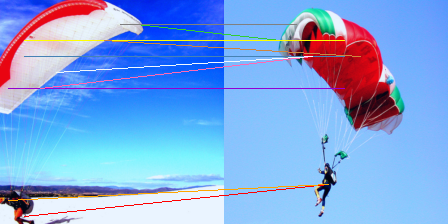} 
    \\
      \includegraphics[width=0.2\linewidth]{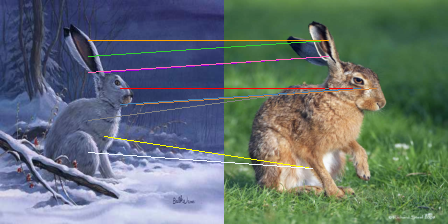} 
    & \includegraphics[width=0.2\linewidth]{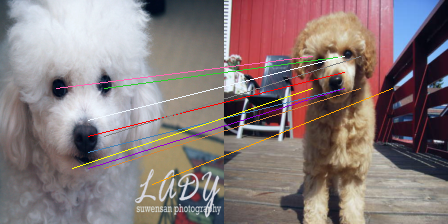} 
    & \includegraphics[width=0.2\linewidth]{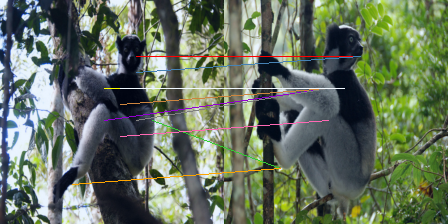} 
    & \includegraphics[width=0.2\linewidth]{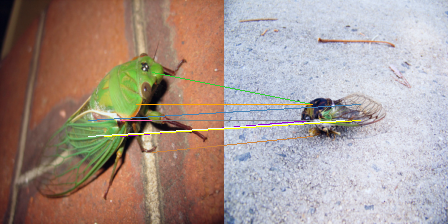} 
    \\
      \includegraphics[width=0.2\linewidth]{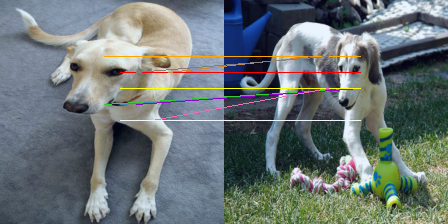} 
    & \includegraphics[width=0.2\linewidth]{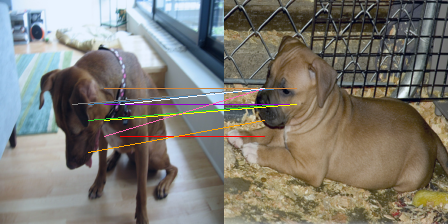} 
    & \includegraphics[width=0.2\linewidth]{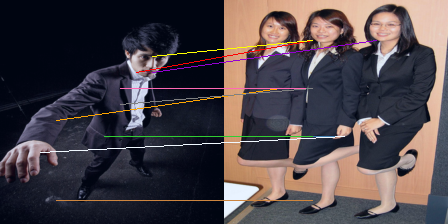} 
    & \includegraphics[width=0.2\linewidth]{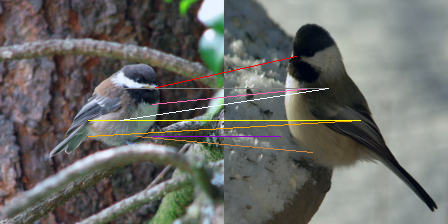} 
    \\

    \end{tabular}
    }
    \vspace{-1em}
    \caption{Patch correspondences of our proposed MGC on MS COCO2017. 
    The given pairs are composed of different images with the same categories.
    }
    \vspace{-1em}
    \label{fig:corresp_class}
\end{figure*}

\section{More Visualization Results of Patch Correspondences}

To further validate the localization capability of our method, we provide more visualization results of patch correspondences on MS COCO2017 validation set. 
For the patch correspondences between different views from the same image instance, the visualization results are presented in Figure~\ref{fig:corresp_instance}. 
The results illustrate that our method can well localize the matching parts of the given positive views, even when the appearance variance is large. 
For the patch correspondences between different image instances with the same categories, the visualization results are presented in Figure~\ref{fig:corresp_class}. 
Our method effectively matches the corresponding patches from the same categories of different images. 
The above visualization results indeed support the representations of our method possess the excellent localization capability.

\end{document}